\documentclass[conference]{IEEEtran} 
\IEEEoverridecommandlockouts

\usepackage{multirow}
\usepackage{algorithm}
\usepackage{algpseudocode}
\usepackage{amssymb}

\usepackage{cite}

\usepackage{graphicx}
\usepackage{amsmath}
\usepackage{xcolor}
\usepackage{algpseudocode}
\usepackage{array}
\usepackage{url}
\usepackage{subcaption}
\usepackage{hyperref}

\title{D3VL: Understanding Driving Scenes from 3D Time Series Data and Video with Language Models}

\author{Heesang Han\textsuperscript{\rm 1, 3}, A. Lynn Abbott\textsuperscript{\rm 1,3}, Abhijit Sarkar\textsuperscript{\rm 1,2,3} \\
\textsuperscript{\rm 1}Bradley Department of Electrical and Computer Engineering, Virginia Tech, USA \\
\textsuperscript{\rm 2}Virginia Tech Transportation Institute, USA \\
\textsuperscript{\rm 3}Sanghani Center for Artificial Intelligence and Data Analytics, USA \\
{\tt\small \{heesang, abbott, asarkar1\}@vt.edu}
}

\begin{document}
\maketitle

\begin{abstract}

Recent advances in Multimodal Large Language Models (MLLMs) have triggered the development of end-to-end MLLMs for autonomous driving. However, the main emphasis to date has been for MLLMs using 2D images and videos. In contrast, this paper considers MLLM effectiveness using 3D sensors, particularly LiDAR and stereo cameras. LiDAR presents unique challenges to integration within an MLLM, largely because of data sparsity and lack of a grid structure for the data.
For similar reasons, fusion of camera and LiDAR data within an MLLM pipeline is also uncommon. However, most autonomous systems rely on LiDAR-based sensing, and incorporating 3D data has been proven to improve performance in traditional 3D scene perception tasks. This paper presents \textit{D3VL}, a novel MLLM framework that integrates 2D and 3D time-series data in a single but simple architecture. The model aims to answer questions involving traffic scene understanding and safety. \textit{D3VL} shows an 11\% improvement in the KITTI Question-Answering (QA) dataset compared to baseline methods in processing 2D and 3D time-series data. This paper further introduces the Waymo QA dataset extension, which assesses models' capabilities in processing 3D and time-series data under diverse driving conditions. \textit{D3VL} implementation code and WaymoQA extension can be found on our supplemental website: \url{https://automotivesafety-lvlm.github.io}.

\end{abstract}

\section{Introduction}

\begin{figure*}[!t]
    \centering
    \vspace{5pt}
    \includegraphics[width=0.95\linewidth]{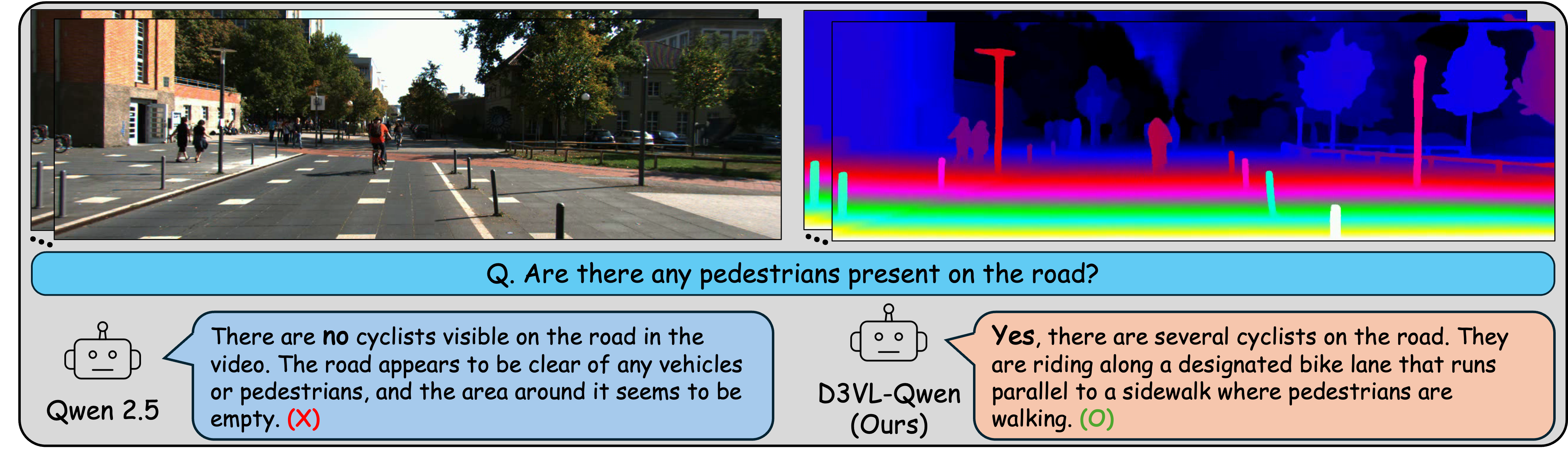} \\
    \caption{Examples in processing 2D and 3D time-series data. \textit{D3VL} framework can effectively utilize 3D data when it is difficult to find smaller cyclists from 2D image, compared to baseline Qwen 2.5 model. More examples are available in our website.}
    \label{fig:teaser}
    \vspace{-0.1in}
\end{figure*}

Recent advances in Multimodal Large Language Models (MLLM) have revolutionized Intelligent Transportation Systems (ITS) \cite{wandelt_large_2024}, excelling in complex reasoning tasks essential for ITS operations. Further, researchers have shown promising results in designing MLLMs for autonomous driving \cite{hwang_emma_2025, sima_drivelm_2024, zhou_autovla_2025}. MLLMs have partially addressed the long-tailed object detection problem with their vast pretrained knowledge \cite{madan_revisiting_2024, han_few-shot_2024}, showing potential to help AV systems detect rare objects such as emergency vehicles and traffic cones. MLLMs also excel in adapting to different domains, unlike traditional systems, which falter when deployed in different countries \cite{li_driving_2024}.

However, it is unknown whether current MLLMs could answer safety related questions related to 3-Dimensional (3D) data scene perception. MLLMs' cannot perceive 3D temporal data direct from LiDAR and stereo cameras, hence it cannot infer the dynamics of the scene from 3D data. One main reason for this limitation is practical: constructing and annotating large traffic datasets is difficult, and processing 3D and temporal data generally requires specialized models. In addition, MLLMs require extensive testing with multiple tools, such as benchmarking or visualizers, before being deployed to critical applications such as Autonomous Vehicles (AV). 

\begin{figure*}[ht!]
    \centering   
    \begin{subfigure}[b]{0.3\textwidth}
        \centering
        \includegraphics[width=\textwidth]{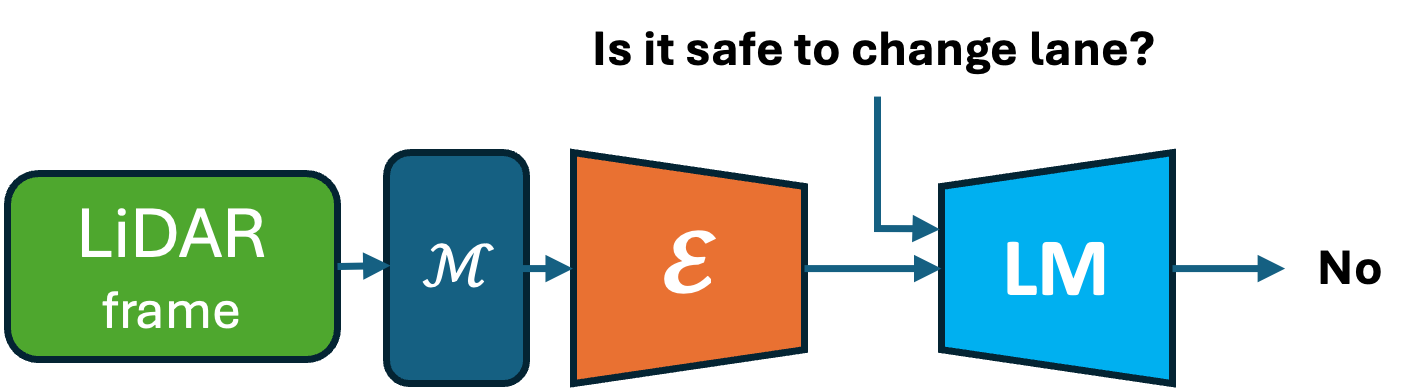}
        \caption{}
        \label{fig:sub1}
    \end{subfigure}
    \hfill
    \begin{subfigure}[b]{0.3\textwidth}
        \centering
        \includegraphics[width=\textwidth]{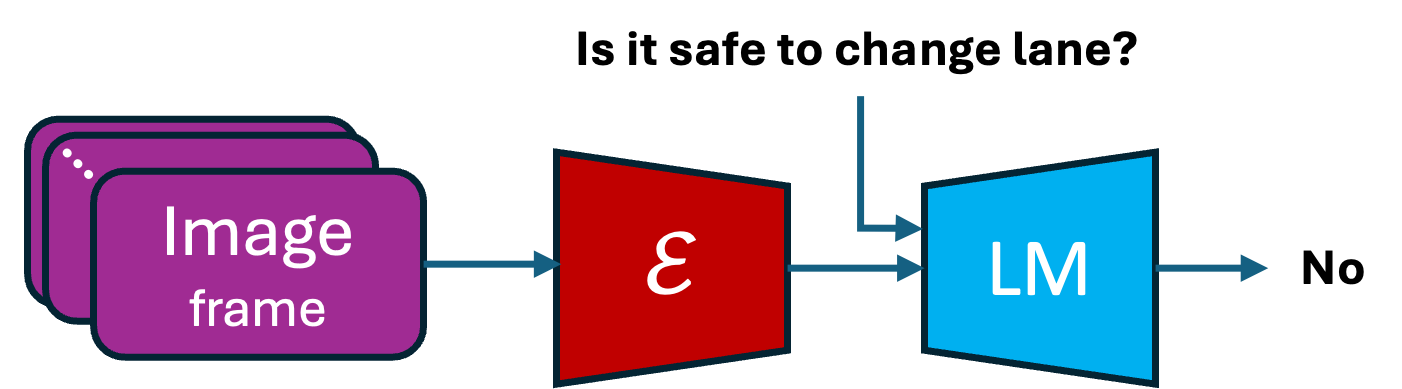}
        \caption{}
        \label{fig:sub2}
    \end{subfigure}
    \hfill
    \begin{subfigure}[b]{0.38\textwidth}
        \centering
        \includegraphics[width=\textwidth]{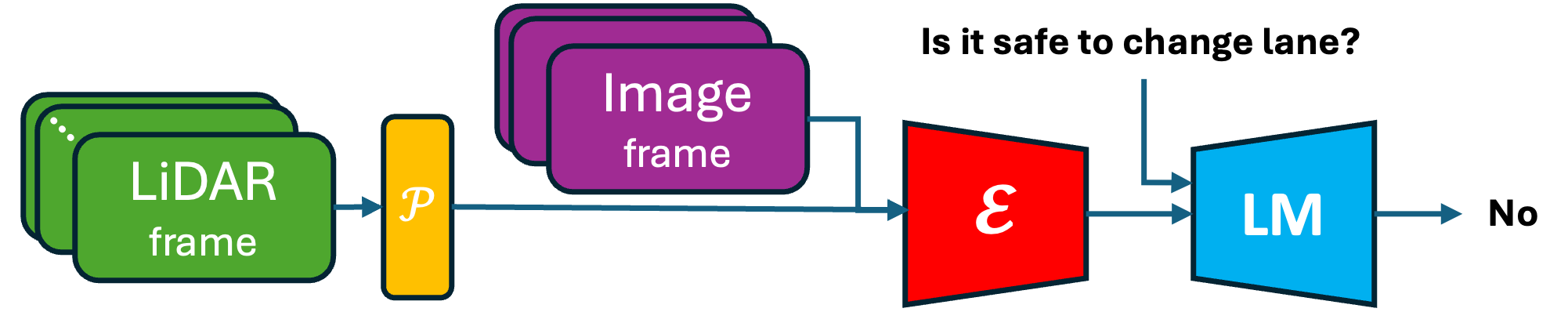}
        \caption{}
        \label{fig:sub3}
    \end{subfigure}

    \caption{Previous MLLM systems for AV applications have used either a) a single LiDAR frame with additional learnable LiDAR processing units $M$ (\cite{yang_lidar-llm_2025, qian_nuscenes-qa_2024, zhi_lscenellm_2025}), or b) multiple image frames (\cite{hwang_emma_2025, xing_openemma_2025, marcu_lingoqa_2024, sima_drivelm_2024}).
    The $P$ module denotes LiDAR-to-camera projection, $\epsilon$ represents a vision encoder, and $LM$ refers to Language Model. Proposed \textit{D3VL} method incorporates both temporal LiDAR data and camera data, and does not require an additional learnable 
    module for LiDAR processing.}
    \label{fig:intro}
    \vspace{-0.15in}
\end{figure*}

Prepending trainable modules to pretrained LLM is a popular way to add a new modality. Recent image-centric MLLMs \cite{lu_deepseek-vl_2024, bai_qwen25-vl_2025, liu_visual_2023} integrate vision encoders and vision-language aligners with pretrained LLMs. MLLMs with LiDAR modalities \cite{yang_lidar-llm_2025, huang_chat-scene_2024} were built by adding pointcloud encoder and pointcloud-text aligner. MLLM's video understanding capabilities could be enhanced by adding video processing modules to image-centric MLLMs. \cite{zhang_videollama_2025, cheng_videollama_2024, zhang_video-llama_2023, zhang_llava-video_2025}. These methods benefit from efficient finetuning thanks to pretrained LLMs, enabling smaller models to perform comparably to large, task-agnostic MLLMs. However, such trainable modules often require complex finetuning strategies and increase model size. 

Multimodal Time-series data processing is also essential for development and understanding of autonomous driving; it requires the tracking and motion prediction of vehicles, pedestrians, and other moving objects near the ego vehicle, and it cannot be achieved without temporal data processing. Jain et al. \cite{jain_semantic_2024} has shown that providing tracking information to MLLMs improves performance in question-answering in the context of AV. Furthermore, numerous studies have shown that sensor fusion of 2D Cameras and 3D LiDAR improves the performance of AV perception models \cite{liu_bevfusion_2023} in various tasks such as 3D object detection; yet, MLLMs' capability in processing temporal LiDAR and camera data has not been explored.

To enable End-to-End (E2E) MLLMs in autonomous driving, they must be compact and simple for onboard units while ensuring top safety performance. This paper investigates the ability of MLLMs in processing temporal LiDAR and camera data and introduces \textit{D3VL} - Driving Scene Understanding with 3D Time-Series Data and Video with Language Model - a novel framework for existing MLLMs to perform better in understanding 2D and 3D time-series driving scenes. Experiments show that the proposed \textit{D3VL} framework improves the baseline MLLMs' accuracy by 11\% when tested with the KITTI QA dataset \cite{geiger_vision_2013, jain_semantic_2024}.
Key contributions include: 

\begin{itemize}
    \item This paper proposes \textit{D3VL}, a novel framework to enhance the performance of smaller MLLMs in processing LiDAR and camera time series data through fine-tuning.
    
    \item This paper introduces WaymoQA, a Video Question-Answering (VideoQA) extension for the Waymo Open Dataset (WOD) \cite{sun_scalability_2020}, focusing on scene understanding and AV safety. 

\end{itemize}


\section{Related Work}

\subsection{Foundational Language Models and MLLMs}

Recent advances in MLLMs began with contrastive learning on text-image pairs \cite{radford_learning_2021, li_blip_2022, alayrac_flamingo_2022, zhai_sigmoid_2023}, learning the relationship between them. Inspired by early LLMs \cite{devlin_bert_2019}, MaskVLM \cite{kwon_masked_2023} added masked vision-language learning to the contrastive learning. Recent MLLMs \cite{liu_visual_2023, bai_qwen25-vl_2025, lu_deepseek-vl_2024} use a pretrained LLM backbone, a vision encoder, and a vision-language aligner; then they are finetuned in multiple stages. Instead of an aligner, LLaMA 3.2-VL \cite{grattafiori_llama_2024} alternates self-attention layers from the LLM backbone with cross-modal attention layers. Newer models \cite{metaai_introducing_2024} adopt Mixture-of-Experts with early fusion to pretrain on large unlabeled data. State-of-the-art closed-source MLLMs \cite{openai_gpt-4_2024, comanici_gemini_2025} often match or surpass open-source models across many tasks.

\begin{figure*}[!t]
    \centering
    \vspace{5pt}
    \includegraphics[width=0.95\linewidth]{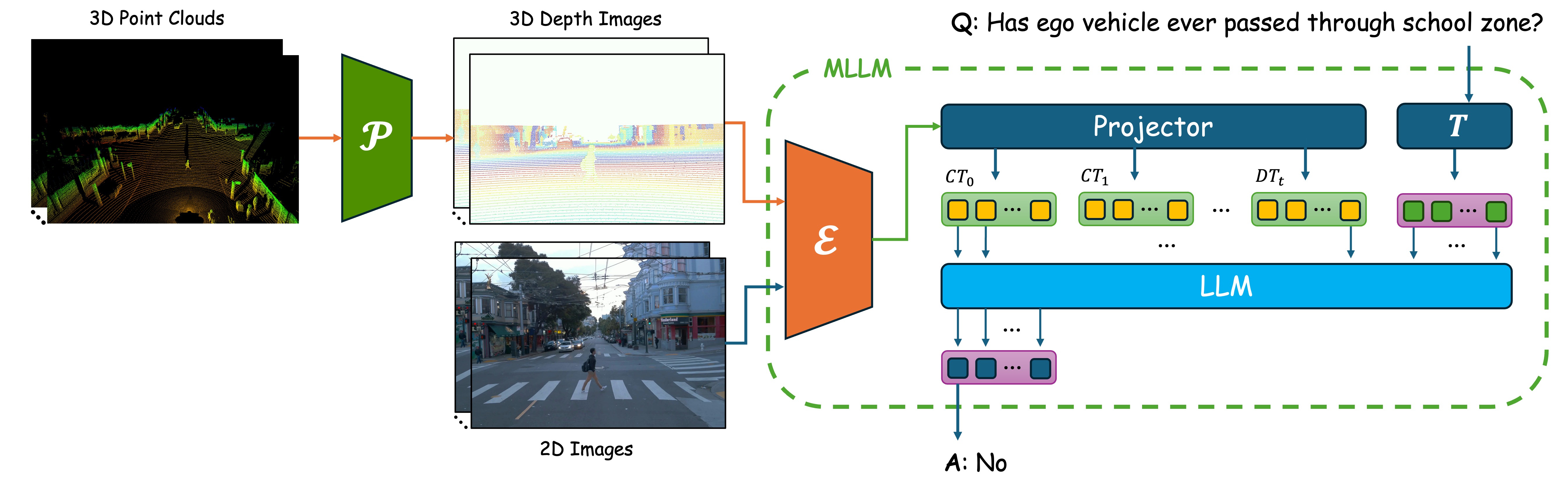}
    \vspace{-0.1in}
    \caption{Overall architecture for \textit{D3VL}. $\mathcal{P}$ is LiDAR-to-Camera projection as in algorithm~\ref{alg:1}. $\varepsilon$ is a transformer vision encoder in an MLLM, projector is vision-language projector, $T$ is text tokenizer, and LLM is foundational LLM decoder, such as Qwen-2.5.}
    \label{fig:arch}
    \vspace{-0.15in}
\end{figure*}


\subsection{MLLMs for Pointcloud and Video}

Researchers have been enhancing foundational MLLMs to incorporate additional modalities. Inspired by CLIP \cite{radford_learning_2021}, CLIP$^2$ \cite{zeng_clip2_2023} uses contrastive text-image-point pretraining. Chat-3D \cite{wang_data-efficiently_2025} combines a point cloud encoder, projector, relation modules, and an LLM to understand 3D relations in a scene. PointLLM \cite{xu_pointllm_2024} adds a 3D encoder and projector to support LiDAR, while LiDAR-LLM \cite{yang_lidar-llm_2025} uses a bird's-eye-view-aware transformer as a projector instead. LSceneLLM \cite{zhi_lscenellm_2025} applies adaptive scaling for larger scenes, and SpatialBot \cite{cai_spatialbot_2025} is built on prompt engineering with depth images. Although these methods integrate LiDAR with foundational LLMs, they require specialized modules for 3D input that increase model size and complexity. Also, they mostly operate on single LiDAR frames, leaving temporal sequence integration largely unexplored. 

Recent studies \cite{zhang_video-llama_2023, cheng_videollama_2024, zhang_videollama_2025, zhang_llava-video_2025} highlight the potential of integrating video into foundational LLMs. These models concatenate tokens from video frames and feed them to pretrained LLMs, which are then finetuned with instruction or reinforcement tuning so that the model can interpret tokens from a sequence of frames. 

\subsection{MLLMs for Autonomous Driving}
Recent advances in MLLMs have unlocked the potential of E2E MLLMs in AV. EMMA and OpenEMMA \cite{hwang_emma_2025, xing_openemma_2025} explore the potential of Chain-of-Thought prompting in E2E MLLMs for autonomous driving. AutoVLA \cite{zhou_autovla_2025} explores adaptive reasoning and reinforcement tuning in MLLM-based autonomous driving. These models only take camera images, historical ego poses, and high level text instructions; the training process is self-supervised. 

Question-Answering (QA) datasets and methods have been developed to facilitate research on E2E autonomous driving models. DriveLM-NuScenes \cite{sima_drivelm_2024} provides high-quality QA pairs on perception, prediction, and planning for AV safety. LingoQA \cite{marcu_lingoqa_2024} is targeted for training E2E autonomous driving models. ScVLM \cite{shi_scvlm_2025} presents a MLLM-based approach to answer QA in safety-critical events. However, all these datasets lack 3D modality. 

NuScenesQA \cite{qian_nuscenes-qa_2024} contains a large number of QA pairs with camera and LiDAR data, but its questions mainly target spatial relations. In addition, the dataset lacks temporal information, as all scenearios are 0.5 seconds long. The KITTI QA dataset \cite{geiger_vision_2013, jain_semantic_2024} features high-quality QA pairs on AV scene understanding with 3D object tracking but shares the limitations of the KITTI dataset\cite{geiger_vision_2013}: the dataset is small and was collected in ideal condition - clear weather, daytime, and ego vehicle moving slow. Table \ref{tab:dataset} summarizes these findings.

\begin{table}[]
    \centering
    \caption{QA Datasets for Autonomous Driving}
    
    \begin{tabular}{|l|c|c|c|c|}
        \hline
        \multirow{2}{*}{Dataset} & WaymoQA & KITTI QA & NuScenesQA & LingoQA \\
        & \textbf{(ours)} \cite{sun_scalability_2020} & \cite{jain_semantic_2024}& \cite{qian_nuscenes-qa_2024} & \cite{marcu_lingoqa_2024} \\
        \hline
        $\|Train\|$ & 768 & 21 & 28K & 28K \\ 
        $\|Test\|$  & 150 & 29 & 6K & 100 \\
        $\|VQA\|$ & 60K & 2K & 460K & 421K \\ 
        Scene length & 20s & 3s-1min & 0.5s & 4s \\
        Modality & 2D \& 3D & 2D & 2D \& 3D & 2D\\
        \hline
    \end{tabular}
    \label{tab:dataset}
    \vspace{5pt}
    \raggedright
    
    $\|$Train$\|$ and $\|$Test$\|$ are the number of scenes in each dataset. $\|$VQA$\|$ is the total number of Video-Question-Answer triplets in the dataset. 
    \vspace{-0.2in}
\end{table}

\vspace{-1mm}
\section{Methods}
\subsection{D3VL Framework}

All sensors are assumed to be temporally synchronized. 3D data is obtained from a LiDAR sensor or a stereo camera. The proposed framework first produces a depth image from the 3D data. 

Given a LiDAR point cloud $X = {x_1,\ldots, x_n}$ where $x_i = [x \ y \ z \ 1]^T$, LiDAR-to-Camera transformation matrix $E \in \mathbb{R}^{4 \times 4}$, and a camera matrix $P \in \mathbb{R}^{3 \times 4 }$, a depth image $X' \in \mathbb{R}^{W\times H}$ can be obtained using LiDAR-to-Camera projections \cite{hartley_multiple_2004} as in algorithm~\ref{alg:1}. First, all pixels in a depth image $X'$ are initialized to $d_{max}$, the maximum effective range of a sensor used in a dataset. For every point $x_i$ in LiDAR point cloud $X$, $E$ is applied to transform the LiDAR coordinates into the camera coordinates, and then $P$ is applied to project onto the camera plane. If the projected coordinate $(x, y)$ is within the camera frame, a pixel X'(x, y) in depth image is encoded with the depth value $\|x_i\|$.

\begin{algorithm}[!h]
    \centering
    \caption{LiDAR to Camera Projection}
    \begin{algorithmic}[1]
    \State $X'(i, j) \gets d_{max} \ \forall \ (i, j) \in [0, W] \times [0, H]$ 
    \For{$k = {1, \ldots N}$ }
        \State $ p = PEX_k$ \quad \quad // $p = [x', y', 1] \in R^{3}$
        \State $ x \gets p_1/p_3$,  $y \gets p_2/p_3$
        \If{$(x, y) \in [0, W] \times [0, H]$}
        \State $X'(x, y) = min(\|X_k\|_2, d_{max})$
        \EndIf
    \EndFor
    \end{algorithmic}
    \label{alg:1}
\end{algorithm}

\begin{figure*}[!t]
    \centering
    \vspace{5pt}
    \includegraphics[width=0.95\linewidth]{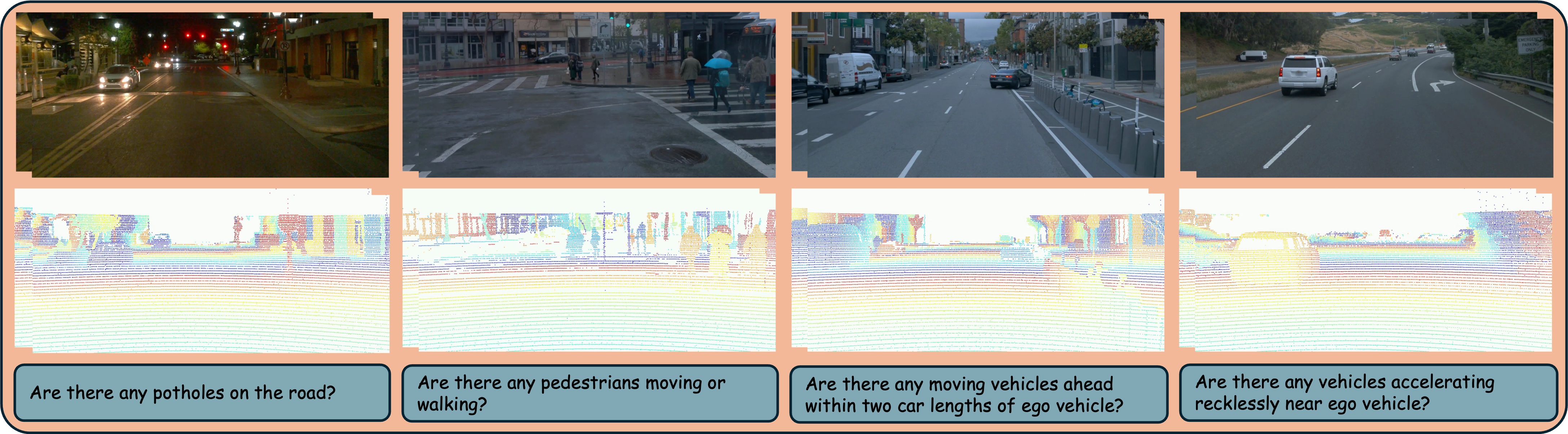} \\
    \caption{Example Video-Question-Answer triplets from Waymo QA dataset extension, showcasing diverse driving conditions and temporal 3D-aware questions. More examples are available in our website.}
    \vspace{-0.15in}
    \label{fig:waymo_teaser}
\end{figure*}
Given a stereo image pair, the framework uses BridgeDepth \cite{guan_bridgedepth_2025} for stereo depth estimation to generate a depth map. Then, a linear colormap is applied to improve visibility. 

At a given time $t$, the framework receives a sequence of $t+1$ camera images $C_{0}, \dots , C_{t}$ and a corresponding sequence of depth images $D_{0}, \dots , D_t$. Each camera image $C_i$ and depth image $D_i$ are tokenized separately to $CT_i$ and $DT_i$ using an image encoder $\varepsilon$ in MLLM. Each token from the 2D image sequence and 3D depth sequence is then projected into 1-D language space with a vision-language projector in MLLM, which is often a multi-layer perceptrons. Text instructions and questions are also converted to tokens $T_{text}$ at this stage. A complete sequence of input tokens $IT$
\begin{equation*}
    IT = Concat(CT_0, \dots, CT_t, \  DT_0, \dots DT_t, \  T_{text})
\end{equation*}
is then created and fed into the LLM to retrieve the final output, which is the answer to the question in this case.  

\textit{D3VL} framework includes a finetuning process to learn rare, AV-specific scenes. During the process, the vision encoder $\varepsilon$ and vision-language projector learn to better encode depth images, while the backbone LLM learns to better decode depth images and rare scenes.  Leveraging the pretrained knowledge of foundational LLMs, \textit{D3VL} only requires at most 3 training epochs on the KITTI QA dataset; no performance improvement has been observed after the 3 epochs. Key training hyperparameters are: learning rate $1\times10^{-4}$, temperature $1.0$, and top-p $1.0$.

\subsection{Dataset}

To the best of our knowledge, this is the first study to explore MLLMs' ability to process timeseries LiDAR and camera data using QA datasets for AV, and no existing QA dataset satisfies all requirements of this study. As shown in Table~\ref{tab:dataset}, some datasets \cite{marcu_lingoqa_2024, sima_drivelm_2024} lack 3D modalities; another \cite{qian_nuscenes-qa_2024, caesar_nuscenes_2020} does not contain meaningful temporal contexts due to short scene length; the other \cite{geiger_vision_2013, jain_semantic_2024} only contains driving scenes captured during clear weather in the daytime. Therefore, this paper introduces WaymoQA, a QA extension for the Waymo Open Dataset (WOD) \cite{sun_scalability_2020}, which includes 60K Video-Question-Answer triplets on autonomous driving. 

WOD features 768 training scenes and 150 testing scenes, each 20 seconds long, showcasing various driving conditions, such as adverse weather, nighttime, rush hours, and construction. Each scene is annotated with 65 QA pairs to assess AV safety in four categories: Environment, Traffic Signals and Signs, Moving Objects, and Planning. These questions are extended from Jain et al.'s work \cite{jain_semantic_2024}, adding various QA pairs that require 3D perception and temporal perception, including questions that require object tracking or speed estimation. The dataset also includes rural scenes with higher vehicle speeds, which are absent from KITTI \cite{geiger_vision_2013} and NuScenes-based \cite{caesar_nuscenes_2020} datasets. All QA pairs use multiple-choice questions to reduce output token generation, which is more expensive than input token processing. Example questions from each area are listed below and in Figure~\ref{fig:waymo_teaser}: 
\begin{itemize}    
    \item \textbf{Environment} \\
        - Are there any potholes on the road? \\     
        - Has ego vehicle passed through a pedestrian crossing? 
        \vspace{5pt}
    \item \textbf{Traffic Signals and Signs} \\
        - What is the last observed speed limit? \\
        - Which light was on when the ego vehicle passed the traffic light? \vspace{5pt}
    \item \textbf{Moving Objects} \\
        - Are there any pedestrians on a road, excluding crosswalk? \\
        - Are there any moving vehicles ahead within two car lengths of ego vehicle?
        \vspace{5pt}
    \item \textbf{Safety \& Planning} \\
        - Is it safe for the ego vehicle to change lanes? \\
        - Are there any vehicles accelerating on the same road as ego vehicle?
\end{itemize}

Human annotation of a dataset is costly. Therefore, WaymoQA is annotated in a semi-supervised fashion, inspired by NuScenesQA\cite{qian_nuscenes-qa_2024}. Human experts annotate testing scenes, and finetuned MLLM annotates the rest. Google Gemini 2.5 Flash \cite{comanici_gemini_2025} underwent supervised and Low-Rank Adapter (LoRA) finetuning with more than 10K of 2D and 3D Video-QA triplets from human-annotated test scenes over 1 epoch using a Rank 8 LoRA. Annotations for training scenes are generated using the finetuned model with Temperature of 0.5 and Top-p of 1.

\begin{table*}[!t]
    \centering
    \vspace{7pt}
    \caption{\textit{D3VL} and baseline MLLMs benchmark on KITTI QA dataset}
    \begin{tabular}{|l|c|c|c|c|c|c|c|c|c|c|c|c|}
        \hline
        \multirow{2}{*}{Methods} & \multicolumn{6}{c|}{2D (Camera)} & \multicolumn{6}{c|}{3D (Stereo) + 2D (Camera)}  \\
        \cline{2-13}
         & RI & VRU & HF & V2VI & OEF & \textbf{Overall} & RI & VRU & HF & V2VI & OEF & \textbf{Overall} \\
        \hline
        \multicolumn{13}{|l|}{\textbf{Baseline methods:}} \\
        \hline
        Qwen-2.5-VL (3B) \cite{bai_qwen25-vl_2025} & 62.17 & 55.56& \textcolor{red}{75.00}& 74.81& 81.48& 68.70 & 37.57& 51.85& 52.31& 61.48& 67.72& 50.93\\
        Qwen-2.5-VL (7B) \cite{bai_qwen25-vl_2025} & 68.25 & \textcolor{red}{64.20}& 51.39& {\color{red}77.78}& {\color{red}91.53}& 69.54& \textcolor{red}{70.63}& 42.59& 71.30& 70.37 & 70.9 & 66.57 \\
        Qwen-2.5-VL (32B) \cite{bai_qwen25-vl_2025} & \textcolor{red}{70.63}& 42.59& 71.30& 70.37& 70.90& 66.57& 68.25& 48.15& 74.07& 58.52 & 63.49 & 64.35 \\
        VideoLLaMA3 (2B) \cite{zhang_videollama_2025} & 38.10 & 38.89& 42.59& 40.74& 47.62& 41.11& 42.33& 36.42& 40.74& 37.04& 49.74& 41.76 \\

        Gemini 2.5 (Flash) \cite{comanici_gemini_2025}& - & - & - & - & - & 71.94 & - & - & - & - & - & \textcolor{red}{74.26} \\
        
        \hline
        \multicolumn{13}{|l|}{\textbf{Existing State-of-the-Art methods:}} \\
        \hline
        Jain et al. \cite{jain_semantic_2024} & 81.56 & \textcolor{blue}{80.43} & 85.60 & 77.78 & 87.11& 81.75& 76.63& 71.53& 89.32& 81.25& 80.65& 79.69  \\
        \hline
        \multicolumn{13}{|l|}{\textbf{\textit{D3VL} (ours): }} \\
        \hline
        \textit{D3VL}-Qwen & 85.19 & 64.20 & \textcolor{blue}{89.81}& \textcolor{blue}{83.70}& 96.83& 84.81& \textcolor{blue}{86.77}& 64.81& 89.35& \textcolor{blue}{83.70}& \textcolor{blue}{97.35} & {\color{blue} 85.46} \\
        \hline
    \end{tabular}
    \label{tab:waymo}
    \vspace{5pt}
    
    \textbf{\textcolor{blue}{Blue}} highlights \textbf{overall} best performance, and \textbf{\textcolor{red}{Red}} highlights best performance \textbf{among baseline methods}. RI, VRU, HFDB, V2VI, and OEF denotes Road Infrastructure, Vulnerable Road User, Human Factors \& Driving Behavior, 
    Vehicle-to-Vehicle Interaction, anb Other Environment Factors, respectively
    \vspace{-0.15in}
    
\end{table*}

\section{Experiments}
\subsection{Setup}
We extensively evaluate popular MLLMs on processing time-series 3D and 2D data from KITTI QA datasets \cite{geiger_vision_2013, jain_semantic_2024} and Waymo QA datasets \cite{sun_scalability_2020}. As baselines, we test Qwen 2.5 VL (3B, 7B, 32B) \cite{bai_qwen25-vl_2025}, VideoLLaMA3 (2B) \cite{zhang_videollama_2025}, and Gemini 2.5 (Flash) \cite{comanici_gemini_2025}. We also evaluate and compare our \textit{D3VL} framework with the baselines and existing state-of-the-art methods. 

Each baseline method is evaluated zero-shot with (1) 2D time-series input only and (2) 2D and 3D time-series inputs from the testing dataset. The \textit{D3VL} framework with Qwen-2.5-VL-3B backbone is finetuned on 2D and 3D time-series inputs from training dataset and evaluated on testing dataset, which remains unseen during tuning. To assess the impact of 3D data, another \textit{D3VL} variant is finetuned and tested without 3D inputs. All metrics are Top-1 accuracy. \textit{D3VL} is finetuned for 1-3 epochs, as more epochs do not significantly improve performance.

\subsection{Results}

Table~\ref{tab:waymo} summarizes the results on KITTI QA dataset; results on the Waymo QA dataset are available on our supplemental website. Most baselines perform better without 3D input, likely because their pretraining data contain very few 3D samples. Even for larger models, performance on 2D input alone remains substantially better than on 2D and 3D input.

The \textit{D3VL} framework outperforms all baselines across all five areas in KITTI QA, achieving an 11\% gain in overall accuracy. It also surpasses Jain et al.\cite{jain_semantic_2024} in overall performance and in all areas except VRU. Compared to \textit{D3VL} finetuned only on 2D inputs, fine-tuning on both 2D and 3D inputs notably improves RI question processing, with particularly large gains for the following questions:
\vspace{3pt}

\begin{itemize}
    \item RI: Are there any visible road dividers or medians? \textbf{+7\%}
    \item RI: Are there proper road signs indicating turns and intersections? \textbf{+11\%}
    \item VRU: Are there any cyclists on the road? \textbf{+7\%}
    \item VRU: Are there any pedestrians present on the road? \textbf{+7\%}
\end{itemize}

\vspace{3pt}
On the other hand, \textit{D3VL} finetuned on 2D inputs alone performs noticeably better in processing one question: 
\begin{itemize}
    \item VRU: Is there a designated pedestrian crossing? \textbf{+7\%}
\end{itemize}

\section{Discussions}

In general, \textit{D3VL} significantly improves MLLMs' capability to process 2D and 3D time series data by 11\% to 27\%, across different areas in KITTI QA dataset, compared to baselines with both 2D and 3D inputs. Table~\ref{tab:waymo} shows that adding 3D input to the baseline model without \textit{D3VL} hinders their ability to process driving scenes and produce correct answers.

It is noticeable that Jain et al.'s work \cite{jain_semantic_2024} with and without 3D data performs substantially better in answering VRU questions than any other baseline methods and \textit{D3VL} framework. VRU questions assess models' ability to process information regarding pedestrians and cyclists, which are relatively common concepts. Based on the GPT-4 model and trained on a large proprietary dataset, it effectively processes common concepts using only images. Notably, it performs better without 3D data, indicating its heavy reliance on camera images when processing VRU questions. Its performance remains unknown in poor weather or at night, which is not included in the KITTI QA dataset. 

Compared to \textit{D3VL} with 2D input alone, \textit{D3VL} with 3D input shows superior performance in perceiving cyclists, pedestrians, and road dividers, all of which contain rich 3D features that differentiate them from other types of objects that exist in driving scenes. Also, as shown in figure~\ref{fig:teaser}, all of those objects are usually very small in 2D images. The only question \textit{D3VL} finetuned without 3D input includes perceiving pedestrian crossings, which is hard to differentiate from regular roads without pedestrian crossings. The \textit{D3VL} framework effectively helps MLLMs learn prominent 3D features.

\section{Conclusion}

This paper is the first to investigate the capability of MLLMs in processing Camera (2D) and LiDAR/Stereo (3D) time-series data in the context of autonomous driving. While baseline MLLMs achieve 51-74\% accuracy in processing such data in the KITTI QA dataset, the proposed \textit{D3VL} framework improves the ability to handle 2D and 3D time-series data, reaching an accuracy of 85\% and surpassing all existing methods. Especially, the \textit{D3VL} framework helps MLLMs learn prominent 3D features, enhancing MLLMs' capability to perceive information regarding pedestrians and cyclists in driving scenes. Additionally, the paper further identifies limitations in existing AV QA datasets and presents Waymo QA dataset extensions that address those problems. 

\section*{Acknowledgment}
The authors sincerely thank Joe Bekiranov and Jin Woo Baik for their help with the human annotation process of the WaymoQA test dataset.

\bibliographystyle{IEEEtran}
\bibliography{bibtex/bib/references}

\end{document}